\documentclass[runningheads]{llncs}
\usepackage[T1]{fontenc}
\usepackage{graphicx}
\usepackage{amsmath,,amssymb,amsfonts,bm}
\usepackage{algorithm}
\usepackage{algpseudocode}
\usepackage{subfigure}
\usepackage{wrapfig,lipsum,booktabs}
\usepackage[misc]{ifsym}
\usepackage{footmisc}

\begin{document}
\title{Learning to Play Text-based Adventure Games with Maximum Entropy Reinforcement Learning}
\toctitle{Learning to Play Text-based Adventure Games with Maximum Entropy Reinforcement Learning}
\titlerunning{Text-based Adventure Games with Maximum Entropy RL}


\author{Weichen Li\inst{1}\Letter \and
Rati Devidze\inst{2} \and
Sophie Fellenz\inst{1} \orcidID{ 0000-0002-5385-3926}}
\tocauthor{Weichen Li, Rati Devidze, Sophie Fellenzs}
\authorrunning{W.Li et al.}
%
\institute{University of Kaiserslautern-Landau, Kaiserslautern, Germany
\email{\{weichen,fellenz\}@cs.uni-kl.de}\\
Max Planck Institute for Software Systems (MPI-SWS), Saarbrücken, Germany
\email{rdevidze@mpi-sws.org}}

\maketitle              
\begin{abstract}
Text-based adventure games are a popular testbed for language based reinforcement learning (RL). In previous work, deep Q-learning is most often used as the learning agent. Q-learning algorithms are difficult to apply to complex real-world domains, for example due to their instability in training. Therefore, we adapt the Soft-Actor-Critic (SAC) algorithm
to the domain of text-based adventure games in this paper. To deal with sparse extrinsic rewards from the environment, we combine the SAC with a potential-based reward shaping technique to provide more informative (dense) reward signals to the RL agent. The SAC method achieves higher scores than the Q-learning methods on many games with only half the
number of training steps. Additionally, the reward shaping technique helps the agent to learn the policy faster and improve the score for some games. Overall, our findings show that the SAC algorithm is a well-suited approach for text-based games. 

\end{abstract}
\section{Introduction}
Language-based interactions are an integral part of our everyday life. Reinforcement learning (RL) is a promising technique for developing autonomous agents used in real-life applications, such as dialog systems.
A game environment can be considered to be a simulated world, where the agent must play their actions by interacting with the environment to achieve some game-specific goal.
In games, we can compare the performance of various RL agents using the final game score. Therefore, text-based adventure games are a useful benchmark for developing language-based agents \cite{hausknecht2020interactive}.

\begin{figure}[t!]
\centering
    \includegraphics[width=\linewidth]{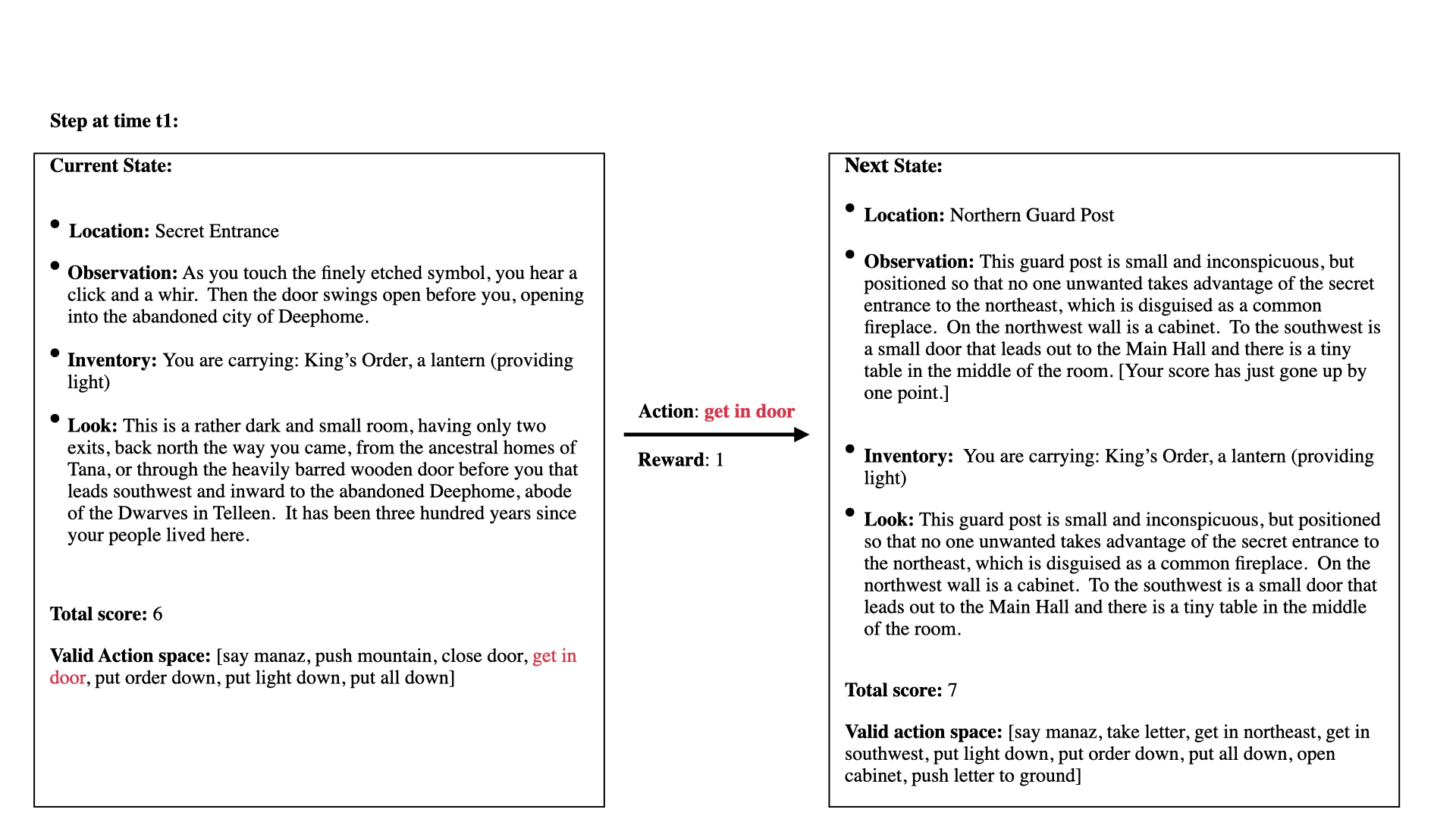}
     \caption{This figure shows one possible state transition for the game \textit{deephome}. At time t1, the RL agent has the current state and valid action space from the Jericho environment. The agent then needs to predict the action (\textit{e.g., \textit{get in door}}) from the valid action space and move to the next state, where it receives a reward from the game environment, and the total score is updated based on the reward. The state information includes location, observation, inventory, and look.}
    \label{fig:example}
\end{figure}

Figure \ref{fig:example} illustrates the problem setup for this paper. The large and discrete action space is the main difference between text-based adventure games and other RL scenarios. In contrast to other games (e.g., ATARI games), each action is characterized by a sentence or word (e.g., \textit{climb tree}). Also, the action space is not fixed. For example, if the agent is in front of the house, the action \textit{open door} is available, whereas if the agent is in the forest, other actions are possible, e.g. \textit{climb tree}, but not \textit{open tree}. Therefore, in addition to the action space, there is the space of valid actions in the current state. This space is much smaller than the space of all actions but can be significantly different in each step. In general, this space of valid actions is unknown to the agent, but a common simplification is to let the agent have the list of valid actions as input and select one from the list.
A number of prior works in this domain focused on the described setup \cite{yao-etal-2020-keep,ammanabrolu2020graph,ammanabrolu2020avoid,guo2020interactive,xu2020deep}. Most of those works used deep Q-learning \cite{mnih2013playing} as a learning agent.

In general, Q-learning has several drawbacks. As an off-policy algorithm, it suffers from high variance, and the performance can be unstable \cite{sutton2018reinforcement}. Other online, policy-based learning algorithms are also unsuitable for our scenario since policy-based agents cannot reuse experiences from the training history. Therefore, in this paper, we develop a learning agent based on the soft actor critic (SAC) algorithm \cite{haarnoja2018soft}, which combines the advantages of both value-based and policy-based learning. It can use a replay buffer as in value-based learning, and it is relatively stable in training as a policy-based algorithm. Additionally, the maximum entropy technique encourages exploration. SAC was originally designed for continuous action spaces; however, with slight modifications, it is applicable for discrete action spaces \cite{christodoulou2019soft}. Nevertheless, it has never been used in the context of text-based adventure games before.

A problem that text-based adventure games have in common with many other RL problems is the sparseness of rewards. Especially at the beginning of training, the agent must perform many actions before receiving feedback. This problem is even more severe in text-based adventure games due to the large and context-dependent action space. To speed up the convergence, it is therefore desirable to have a denser reward function. A popular way to achieve this is through reward shaping. However, 
finding a good reward function is difficult and requires significant manual effort, background information, or expert knowledge. A well-known reward shaping technique, circumventing the need for external knowledge, is potential-based reward shaping \cite{ng1999policy} which has strong theoretical guarantees. This enables faster convergence during the training, as we show for several games.

To sum up, our contributions are as follows:
\begin{enumerate}
    \item We propose to use SAC as an alternative to deep Q-learning for text-based adventure games.
    \item We propose a variant of potential-based reward shaping for discrete action spaces that is effective for text-based adventure games.
    \item We compare our method on a range of games and show that we can achieve better scores than deep Q-learning with fewer training episodes on many games. 
    \item Additionally, we show that convergence is faster with reward shaping for some games.
\end{enumerate}

This paper is structured as follows. We will first review existing approaches to text-based adventure games and potential-based reward shaping in Section \ref{sec:related}. The problem setting and background on soft actor critic RL are introduced in Section \ref{sec:background}. Section \ref{sec:method} discusses our reward shaping method. We present experimental results in Section \ref{sec:results}, discuss the current limitations and future work in Section \ref{limitations},  and conclude in Section \ref{sec:conclusion}.

\section{Related Work}
\label{sec:related}
\textbf{RL for Text-based adventure games}
Hausknecht \textit{et al.} \cite{hausknecht2020interactive} built the Jericho Interactive Fiction environment, which includes 57 different games that are categorized into possible, difficult, and extreme games. 
In general, for text-based adventure games, there are \textit{choice-based} agents and \textit{parser-based} agents \cite{hausknecht2020interactive}. Parser-based agents \cite{narasimhan-etal-2015-language} generate actions using verb-object combinations, whereas choice-based agents choose an action from a pre-generated list of actions. Other related work focuses on action generation \cite{ammanabrolu2020graph,yao-etal-2020-keep,ammanabrolu2020avoid,xu2020deep,guo2020interactive}.
In this work, we follow the line of choice-based agents, which allows us to concentrate on the RL part of our method. 

We compare our experimental results with the deep reinforcement relevance network  (DRRN) \cite{he2015deep} agent. 
DRRN is a widely used framework for choice-based and parser-based agents. The basic idea behind DRRN is to encode the actions and states into embedding vectors separately and then use the state and its corresponding action embeddings as inputs into a neural network to approximate the Q-values of all possible actions $Q(s_t,a_t^{i})$. The action at each time step is selected by $a_t = {argmax}_{a_t^i}(Q(s_t,a_t^{i}))$.

NAIL \cite{hausknecht2019nail} is an agent that is not choice-based, able to play any unseen text-based game without training or repeated interaction and without receiving a list of valid actions.  We compare both DRRN (and variants) and NAIL in our experiments, but only DRRN has the same experimental setup and handicaps as our agent. NAIL serves as a baseline of scores possible without any simplifications of gameplay.

Yao \textit{et al.} \cite{yao2021reading} investigate whether the RL agent can make a decision without any semantic understanding. They evaluate three variants based on DRRN: a) only location information is available as observation b) observations and actions are hashed instead of using the pure text c) inverse dynamic loss based vector representations are used. Their results show that the RL agent can achieve high scores in some cases, even without language semantics. 
In concurrent work, building on this, Gu \textit{et al.} \cite{gu2022revisiting} point out that the RL agent can achieve higher scores by combining semantic and non-semantic representations.
Moreover, Tuyls \textit{et al.} \cite{tuyls2022multi} propose a new framework that includes two stages: the exploitation phase and the exploration phase. The exploitation policy uses imitation learning to select the action based on previous trajectories. The goals of the second exploration policy are to explore the actions to find rewards and reach new states. This work manually adds relevant actions into the valid action space.

The \textbf{potential-based reward shaping} technique that we use was introduced in the seminal work by Ng \textit{et al.} \cite{ng1999policy}. Potential-based reward shaping (PBRS) is one of the most well-studied reward design techniques. The shaped reward function is obtained by modifying the reward using a state-dependent potential function. The technique preserves a strong invariance property: a policy $\pi$ is optimal under shaped reward \emph{iff} it is optimal under extrinsic reward. 
Furthermore, when using the optimal value function $V^*$ under the original reward function as the potential function, the shaped rewards achieve the maximum possible informativeness. 

In a large number of prior studies interested in PBRS,
Wiewiora \textit{et al.} \cite{wiewiora2003principled} propose the \textit{state-action potential advice} methods, which not only can estimate a good or bad state, but also can advise action. Grzes \textit{et al.} \cite{grzes2010online} evaluate the idea of using the online learned abstract value function as a potential function. Moreover,  Harutyunyan \textit{et al.} \cite{harutyunyan2015expressing} introduce an arbitrary reward function by learning a secondary Q-function. They consider the difference between sampled next state-action value and the expected next state-action value as dynamic advice. In our work we focus on the classical algorithm by Ng \textit{et al.} \cite{ng1999policy} as a robust baseline.

\textbf{Rewards in NLP-based RL agents} One of the challenges of using RL to solve natural language processing (NLP) tasks is the difficulty of designing reward functions. There could be more than one factor that affects the rewards, such as semantic understanding and grammatical correctness. Li \textit{et al.} \cite{li2016deep} define reward considering three factors: ``ease of answering'', ``information flow'', and ``semantic coherence'' for dialogue generation tasks.  Reward shaping techniques have also been used in other NLP-based RL tasks; for example,  Lin \textit{et al.} \cite{lin2018multi} use knowledge-based reward shaping for a multi-hop knowledge graph reasoning task. 
The core difference to our model is that we do not pre-define any function or knowledge as a reward signal, instead shaping the rewards automatically.

\section{Problem Setting and Background}
\label{sec:background}

An environment is defined as a Markov Decision Process (MDP) $M := (\mathcal{S},\mathcal{A},T,\gamma,
\\R)$, where the set of states and actions are denoted by $\mathcal{S}$ and $\mathcal{A}$ respectively. $T: \mathcal{S} \times \mathcal{S} \times \mathcal{A} \rightarrow [0,1]$ captures the state transition dynamics, i.e., $T(s' \mid s,a)$ denotes the probability of landing in state $s'$.

The reward $R$ and terminal signal $d$ come from the game environment, and $\gamma$ is the discount factor. The stochastic policy $\pi: \mathcal{S} \rightarrow \Delta (\mathcal{A})$ is a mapping from a state to a probability distribution over actions, i.e., $\sum_a \pi(a|s) = 1$, parameterized by a neural network.

Notice that the valid action space size is variable at each time step. Following Hausknecht \textit{et al.} \cite{hausknecht2020interactive}, we differentiate between game state $s$ and observation $o$, where the observation refers only to the text that is output by the game whereas the state corresponds to the locations of players, items, monsters, etc. Our agent only knows the observations and not the complete game state.

\subsection{SAC for Discrete Action Spaces}
Soft-actor-critic \cite{haarnoja2018soft} combines both advantages of value-based and policy-based learning. The drawback of value-based learning like deep Q-learning is the instability during training where the policy can have high variance \cite{sutton2018reinforcement}. The SAC algorithm includes three elements. The first element are separate predict and critic neural networks, the second is that offline learning can reuse the past collections via replay buffer, which is the same as in deep Q-learning, and the third is that the entropy of the policy is maximized to encourage exploration. The optimal policy aims to find the highest expected rewards and maximize the entropy term $\mathcal{H}(\pi(.|s_t))$:
\begin{equation*}
    \begin{aligned}
        \pi^{\star} =   \underset{\pi}{\arg\max} \sum_{t=0}^T  \mathbb{E}_{(s_t,a_t) \sim \rho_\pi}
        [& r(s_t,a_t) +  \alpha \mathcal{H}(\pi(.|s_t))]
    \end{aligned}
\end{equation*}
where $s_t$ and $a_t$ denote the state and action at time step $t$ and $\rho$ denotes the state-action marginals of the trajectory distribution induced by a policy $\pi$. The temperature parameter $\alpha$ controls the degree of exploration.
The original SAC is evaluated on several continuous control benchmarks. Since we are dealing with discrete text data, we base our method on the framework for discrete action spaces by Christodoulou \cite{christodoulou2019soft}. The key difference between continuous and discrete action spaces is the computation of the action distribution. For discrete action spaces, it is necessary to compute the probability of each action in the action space. The actor policy is changed from $\pi_{\phi}(a_t|s_t)$, a distribution over the continuous action space, to $\pi_{\phi}(s_t)$, a discrete distribution over the discrete action space. 

The SAC algorithm has a separate predictor (actor) and critic. In the following, we first describe the two crucial equations for updating the critic and then the actor policy update.

In the critic part, the targets for the Q-functions are computed by
\begin{equation}
\begin{split}
     y(r,s',d) = & r +\gamma (1-d)
    \left(\min_{i=1,2}\left(Q_{\hat\theta_i}(s')\right)-\alpha  \log\left(\pi_\phi(s'_t)\right)\right),
\end{split}
\label{eq:target_q}
\end{equation}
where in our scenario, the target Q-values and the policy distribution range over the set of valid actions $A_{valid}(s')$ \cite{hausknecht2020interactive}. We use two Q-functions $Q_{\theta_i}$ and two Q target functions $Q_{\hat\theta_i}$, and $i\in\{1,2\}$ is the index of the Q-neural networks. $\gamma \in (0,1]$ is a discount factor, and $d\in \{0,1\}$ is 1 if the terminal state has been reached.

The critic learns to minimize the distance between the target soft Q-function and the Q-approximation with stochastic gradients:  
\begin{equation}
    \begin{aligned}
        &\nabla J_Q(\theta) =
        \nabla \mathbb{E}_{a\sim \pi(s),s\sim D}\frac{1}{B} \sum_{i=1,2} \left(Q_{\theta_i}(s)-y(r,s',d)\right)^2,
    \label{eq:q_approx}
    \end{aligned}
\end{equation}
where $D$ is the replay buffer, and $B$ is the size of mini-batch sampled from $D$. If using double Q-functions, the agent should learn the loss functions of both Q-neural networks with parameters $\theta_1$ and $\theta_2$.

The update of the actor policy is given by:
\begin{equation}
    \begin{aligned}
       & \nabla J_{\pi}(\phi) = \nabla\mathbb{E}_{s\sim D} \frac{1}{B}\left[\pi_t(s)^T[\alpha \log\pi_{\phi}(s) - \min_{i=1,2}(Q_{\theta_i}(s)]\right].
    \end{aligned}
\label{eq:actor_policy}
\end{equation}
where $Q_{\theta_i}(s)$ denotes the actor value by the Q-function (critic policy), and $\log\pi_{\phi}(s)$ and $\pi_t(s)$ are the expected entropy and probability estimate by the actor policy.  

As shown in Algorithm \ref{alg:sac} in lines 8 and 9, Equations \ref{eq:q_approx} and \ref{eq:actor_policy} constitute the basic SAC algorithm without reward shaping, where critic and actor are updated in turn. In the next section, we will explain the reward shaping in lines 2--7 of the algorithm.

\section{Reward Shaping Method}
\label{sec:method}
The original SAC equation is given in Equation \ref{eq:target_q}. In the following we describe how we are modifying it through reward shaping. The whole algorithm is given by Algorithm \ref{alg:sac}. We start by reward shaping in line 2. The shaping reward function $F: S\times A\times S\rightarrow \mathbb{R}$ \cite{ng1999policy} is given by
\begin{equation}
\label{eq:PBRS_general_form}
F(s,a,s') = \gamma 	\Phi(s') - \Phi(s),
\end{equation}
where $s'$ is the target state and $s$ refers to the source state. Two critical criteria of a reward function are \textit{informativeness} and \textit{sparseness} \cite{devidze2021explicable}. As mentioned in Section \ref{sec:related}, when using the optimal value-function $V^*$ under original reward  as the potential function, i.e., $\Phi(s)=V^*(s)$ , the shaped rewards achieve the maximum possible informativeness.

\begin{algorithm*}
\caption{SAC with potential-based reward shaping}\label{alg:sac}
\begin{algorithmic}[1]
\Require policy $\pi$; Q-functions  $\theta_1,\theta_2, \hat{\theta_1},\hat{\theta_2}$; replay buffer D; roll-out N
\For{step $=1\hdots$ max step}
\newline
\Comment{Update the \textit{critic}:}
    \If{Reward Shaping is True}
        \State{$V_{step}(s) \leftarrow\pi (s)^T\left[(Q_{\hat\theta_i}(s)-\alpha \log(\pi(s))\right)]$ (Eq. \ref{eq:soft_state_value})}  \Comment{Compute soft state val.}
        \State $V_{\text{step}}(s) \leftarrow (1-\alpha)V_{\text{step}}(s) + \alpha (r + \gamma_r V_{\text{step}}(s'))$ (Eq.\ref{eq:rsv}) \Comment{Update value func.}

        \State $F_{\text{step}}(s,a,s') \leftarrow \gamma_r V_{\text{step}}(s') - V_{\text{step}}(s)$ (Eq. \ref{eq:f})  \Comment{Compute shaping function}
        \State $\hat{R}(s,a) \leftarrow R(s,a) + F_{\text{step}}(s,a,s')$ (Eq. \ref{eq:reward}) \Comment{Compute reshaped reward}
    \EndIf
\State{Update Q-function (Equation \ref{eq:q_approx})}
\newline
\Comment{Update the \textit{actor}:}
\State{Update policy (Equation \ref{eq:actor_policy})}
\EndFor
\end{algorithmic}
\end{algorithm*}

Since we do not have access to the optimal value function $V^*$,  we use the idea of \textit{dynamic} reward shaping. In particular, Grzes \textit{et al.} \cite{grzes2010online} generalized the form in Equation \ref{eq:PBRS_general_form}
to dynamic potentials,  and
empirically showed an advantage in helping the agent. 
The idea is that the RL agent uses the current approximation of the value function as a potential function. More precisely, the shaped function $F$ at learning step can be represented as follows (Algorithm \ref{alg:sac}, line 5): 
\begin{equation}
 F(s,a,s') = \gamma_r V(s') - V(s),
 \label{eq:f}
\end{equation}
where $\Phi(s)$ from Equation \ref{eq:f} is given by $V(s)$.
Hence, the new shaped reward $\hat{R}: A\times S\rightarrow \mathbb{R}$ is defined as:
\begin{equation}
    \hat{R}(s, a) := R(s,a) + F(s,a,s'),
    \label{eq:reward}
\end{equation}
where $R(s, a)$ is the original extrinsic reward from the environment (Algorithm \ref{alg:sac}, line 6).

To shape reward signals, we use the soft state value function instead of the plain value function. This allows us to use reward shaping without a separate neural network for the reward function. Haarnoja \textit{et al.} \cite{haarnoja2018soft} also mention that it is in principle not necessary to add a separate approximator for the state value although they find it to stabilize results in practice. More precisely, we directly utilize the original form of the soft value function as given in the SAC algorithm for discrete action spaces \cite{christodoulou2019soft}:

\begin{equation}
V(s) =\pi (s)^T\left[(Q_{\hat\theta_i}(s)-\alpha \log(\pi(s))\right)],
\label{eq:soft_state_value}
\end{equation}
where $Q$ denotes the target Q-functions. 
The soft value has two terms, the expected Q-value at the given state and the entropy regularized probability of all possible actions.
The Q-function aims to update the policy to maximize the expected reward. The maximum entropy policy encourages the agent to explore states with more uncertainty under the given constraints  \cite{ziebart2010modeling}. 

Using Equation \ref{eq:soft_state_value}, inspired by Grze \textit{et al.} \cite{grzes2010online}, the current value $V(s)$ is updated by simple TD learning: 
\begin{equation}
    V(s) = (1-\alpha)V(s) + \alpha (r + \gamma_r V(s')), 
    \label{eq:rsv}
\end{equation}
where, $\alpha$ is the learning rate between 0 and 1, and $\gamma_r$ is the discount rate for TD update.
Now, we can rewrite the target Equation \ref{eq:target_q} by incorporating Equation \ref{eq:f}:
\begin{equation}
    \begin{aligned}
       &y(r,s',d) = [r+(\gamma_r V(s') - V(s))] + \gamma_r (1-d) V(s').
    \end{aligned}
\end{equation}
This concludes the description of our reward shaping algorithm which relies on the soft value function.

\section{Experimental Results}
\label{sec:results}

\subsection{Datasets}
The experiments are run on the Jericho environment
\cite{hausknecht2020interactive}\footnote{https://github.com/microsoft/jericho}, which categorizes the games into three groups: possible games, difficult games, and extreme games. In the following experiments, we focused on the possible and difficult games. The learning agent can solve the possible games in a few steps. The difficult games have sparser rewards and require a higher level of long-term decision-making strategies than the possible games. 

\subsection{Experimental Settings}
We built a choice-based agent \footnote{Source code and additional results of our experiments are available at https://github.com/WeichenLi1223/Text-based-adventure-games-using-SAC \label{footnote_1}}. The agent predicts one of the possible actions from the action space distribution based on the observation of the current time step and the previous action from the last time step. The agent receives the valid action space identified by the world-change detection handicap from the Jericho game environments. Using the same handicaps as the DRRN method, we also use the Load, Save handicap to receive information on inventory and location without changing the game state. As shown in Table \ref{tab:result}, we ran the main experiments using the SAC algorithm. In Figure \ref{fig:results_all} we compare two additional reward shaping variants:
\begin{enumerate}
\item  SAC: This is the basic RL agent with the SAC algorithm.
 \item  SAC+RS: Here we use the reward shaping technique in combination with SAC. This is our SAC with potential-based reward shaping algorithm as given in Algorithm \ref{alg:sac}.
 \item SAC+0.1*RS: This variant is the same as SAC+RS except that it introduces one more parameter $s$ to re-scale the reshape reward, $s * \hat{R}(s, a)$. In the following experiments, we set the value of $s$ as 0.1.
\end{enumerate}

\textbf{Neural networks and parameters} The policy neural network includes three linear layers with two hidden dimensions $D_1 = 512$ and $D_2 = 128$, each hidden layer connects with the ReLU activation function, and the categorical distribution is on top to ensure that the sum of action probabilities is one. The Q-function neural network has also three linear layers with ReLU activation functions. Both policy and Q-function update at each step, and the target Q-functions update the weights from the Q-function every two steps.

\textbf{The RL agent parameters} were set as follows: the batch size is 32, and the learning rate of both policy and Q-function neural networks is 0.0003. The discount factor for SAC update is 0.9. When the value of the discount factor is close to 1, the future rewards are more important to the current state. Epsilon-Greedy action selection and a fixed entropy regularization coefficient were used in all of the experiments. For each game, we ran 8 environments in parallel to get the average score of the last 100 episodes, and each model ran three times to compute the average scores. The maximum number of training steps per episode is 100.
Since the RL agent interacts with the game environments, the training time depends on the game implementation in the Jericho framework. For example, zork1, and zork3 are comparably fast to train, whereas gold takes an extremely long time compared to the rest of the games. Because of this, we only trained gold for 3,000 steps, yomomma for 10,000 steps, and karn for 20,000 steps. Our comparison methods also use varying step sizes for these games (but they use more training steps than we do). Most of the previous work trained the agent in a maximum of 100,000 steps, whereas the maximum number of training steps for our method is only 50,000 in all experiments. 
We use the same parameters for all of the games; however, we recommend trying different values for parameters such as learning rate, a discount factor of reward shaping, or the layers of the neural network for individual games to achieve higher scores.

\textbf{Input representation} Following Hausknecht \textit{et al.} \cite{hausknecht2020interactive}, the state $s$ includes three elements: (observation, inventory, look) at the current time step. The representation of the elements in the state and the action are tokenized by a SentencePiece \cite{kudo2018sentencepiece} model and then use seperate GRUs to learn the embeddings. The embedding size is 128. During training, the agent randomly samples the data from the replay buffer.

\subsection{Results}
We compare our results with the previous choice-based agents using deep Q-learning, and then discuss the effect of reward shaping and its variants.

\setlength{\tabcolsep}{6pt}
\begin{table*}[!htbp] 
\centering
\begin{tabular}{l|rrrr|r||r}
    & &  \multicolumn{3}{c|}{Hausknecht \textit{et al.}  \cite{hausknecht2020interactive}}&\multicolumn{1}{c||}{Yao \textit{et al.} \cite{yao2021reading}}&\multicolumn{1}{|c}{Ours}\\%
\textbf{Game} &Max& \textbf{RAND} & \textbf{DRRN}& \textbf{NAIL}&\textbf{DRRN}&\textbf{SAC} \\
adventureland & 100 & 0 & 20.6 & 0 &- &\textbf{24.8}\\
detective & 360 & 113.7 &197.8& 136.9& \textbf{290} & 274.8 \\
pentari &70 & 0 & 27.2 &0 & 26.5 & \textbf{50.7}\\
 \hline
 \hline
balances & 51 & 10 &10 &10&10& 10 \\
gold    & 100 & 0  & 0 &3 & -& \textbf{6.3}\\
jewel &90  & 0 &1.6 &1.6& - & \textbf{8.8}\\
deephome & 300 & 1  &1 &13.3& \textbf{57} & 48.1 \\
karn & 170 & 0  & \textbf{2.1}&1.2&- & 0.1 \\
ludicorp &150 & 13.2  &13.8 &8.4& 12.7& \textbf{15.1}  \\
zork1 & 350 & 0  & 32.6&10.3&  \textbf{39.4} & 25.7   \\
zork3 & 7  & 0.2  & 0.5&1.8&0.4 & \textbf{3.0} \\
yomomma & 35 & 0 &0.4 &0&- & \textbf{0.99}  \\
\end{tabular}
\caption{\label{tab:result}
The average score of the \textbf{last} 100 episodes is shown for three repetitions of each game. The maximum number of training steps is 50,000 for our method. Top (adventureland, detective, pentari) are possible games and rest of the games are diffifult games.
RAND, DRRN, and NAIL results are by Hausknecht \textit{et al.} \cite{hausknecht2020interactive}, the DRRN in column 6 is the reproduced score by Yao \textit{et al.} \cite{yao2021reading}.}
\end{table*}

\subsubsection{Comparison to Q-learning Methods}
\label{sec:results_q}
Table \ref{tab:result} shows the average game scores
of the SAC-based learning agent on twelve different games over three runs.  All of the twelve different games over three runs. In comparison with DRRN and Yao \textit{et al.} \cite{yao2021reading}, which are deep Q-learning-based RL agents, seven of the SAC agents can achieve notably higher scores while only using half of the training steps. One game (\textit{balances}) got identical scores. The scores of \textit{detective}, \textit{deephome}, \textit{zork1} and \textit{karn} are lower than those using the deep Q-learning agent. The difficult games still include several games where no method has achieved a score higher than a random agent. Same as for the baselines, we compute the average of the last 100 episodes for each run of the game.  For each run of one game, eight environments are run in parallel and the average score is computed. The results of the baselines are taken directly from the respective papers. 

One key idea behind SAC is maximizing the log probability, which can encourage the agent to explore uncertain states and converge faster. The training progress is shown in Figure \ref{fig:results_all} where the game score is plotted over training episodes including standard deviations. We can see that the method converges well except for the games \textit{balance} and \textit{karn} in the additional results, where the agent is not able to learn (see Section \ref{limitations} for a possible explanation). Overall, the results indicate that SAC is well-suited to solve text-based games.
\begin{figure*}[t!]
    \centering
   \subfigure[zork3]{\includegraphics[width=0.3\textwidth]{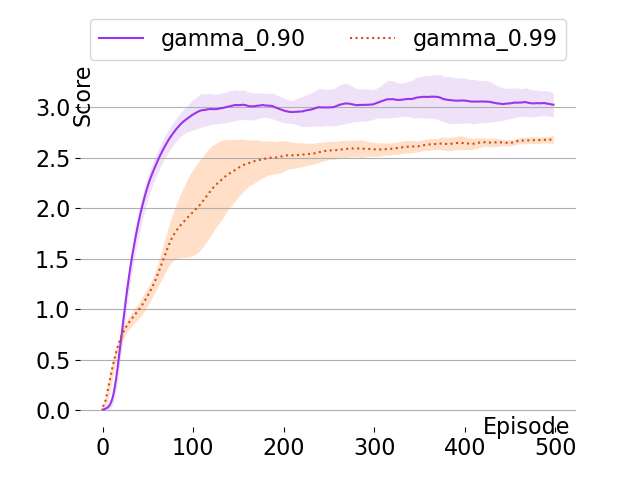}}
   \subfigure[ludicorp]{\includegraphics[width=0.3\textwidth]{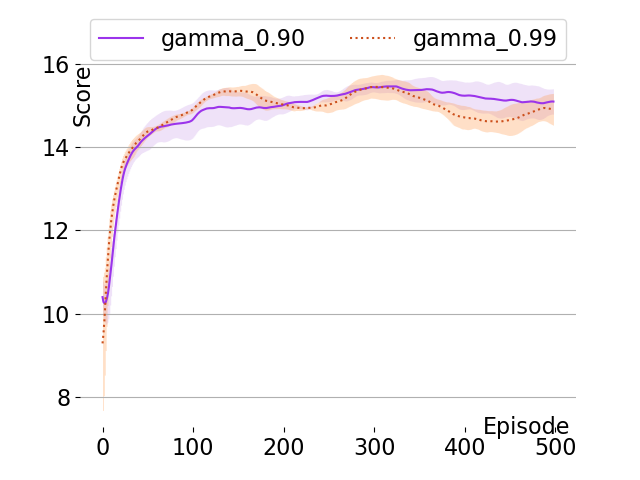}}
    \subfigure[yomomma]{\includegraphics[width=0.3\textwidth]{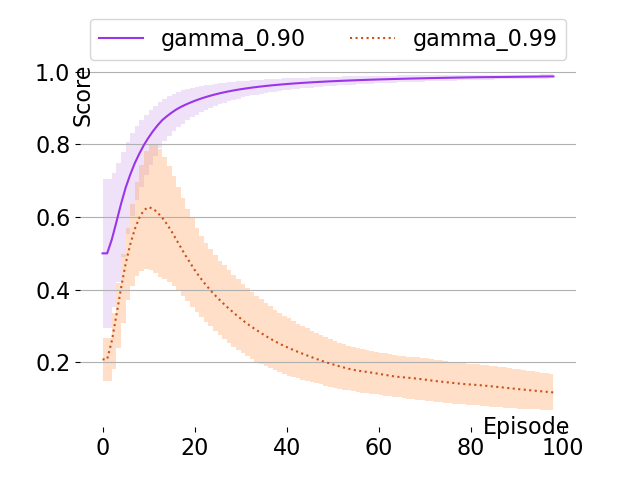}}
    \caption{This figure shows three examples of training the SAC agent with the discount factor of 0.9 and 0.99, where shaded areas correspond to standard deviations. We can see that the agent with a discount factor of 0.9 outperforms the agent with a discount factor of 0.99.}
    \label{fig:ga_sac}
\end{figure*}

Additionally, we observed that the discount factor is a critical parameter during the experiments. As shown in Figure \ref{fig:ga_sac}, using a discount factor 0.9 achieved higher scores than a discount factor of 0.99. When using a discount factor of 0.9 the current update is less dependent on expected future rewards. This might be explained by the highly complex and hard to predict game environments.

\newcommand{\rot}{70}
\setlength{\tabcolsep}{2.5pt}
\begin{table}[t!]
\centering
\begin{tabular}{c|cccccccccccc}
& \rotatebox{\rot}{adventureland} & \rotatebox{\rot}{detective} & \rotatebox{\rot}{pentari} & \rotatebox{\rot}{balances} & \rotatebox{\rot}{gold} & \rotatebox{\rot}{jewel} & \rotatebox{\rot}{deephome} & \rotatebox{\rot}{karn} & \rotatebox{\rot}{yomomma} & \rotatebox{\rot}{ludicorp} & \rotatebox{\rot}{zork1} & \rotatebox{\rot}{zork3} \\
\midrule
SAC & \textbf{24.8} & 274.8 & 50.7& 10 & 6.3 & 8.8 & 48.1 & 0.1 & \textbf{0.99} & \textbf{15.1} & 25.7 & 3.0 \\
RS & 23.8 & 276.2 & \textbf{51.5} & 10 & \textbf{6.7} & \textbf{9.9} & \textbf{50.1} & 0.0 & 0.98 & 14.4 & \textbf{36.0} & 2.8 \\
0.1*RS & 24.6 & \textbf{276.9} & 47.1 & 10 & 6.4 & 9.0 & 49.5 & \textbf{0.2} & 0.98 & 14.6 & 25.1 & \textbf{3.5} \\
\bottomrule
\end{tabular}
\caption{This Table shows that reward shaping improves the performance of SAC for many of the games. Compared are SAC with two variants of reward shaping: non-scaled RS and re-scaled reshaped reward 0.1*RS.}\label{rs-tab:1}
\end{table}

\subsubsection{Comparison to Reward Shaping}
\label{sec:results_reward}

\begin{figure*}[t!]
    \centering
    \subfigure[detective]{\includegraphics[width=0.3\textwidth]{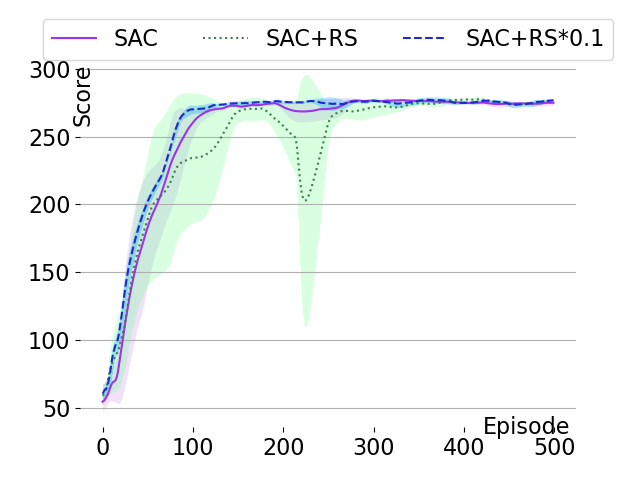}}
    \subfigure[deephome]{\includegraphics[width=0.3\textwidth]{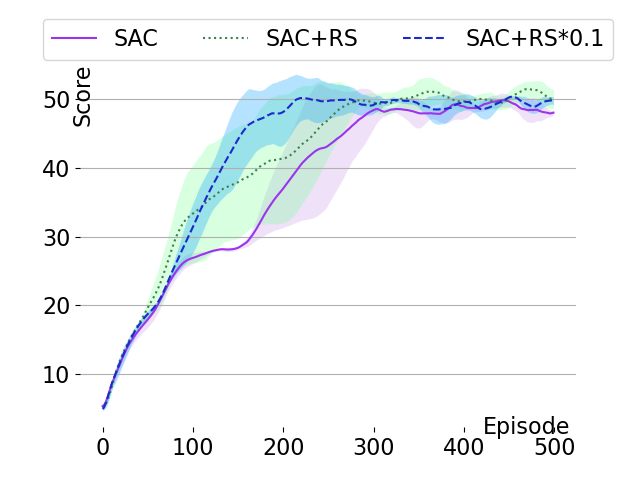}} 
     \subfigure[zork3]{\includegraphics[width=0.3\textwidth]{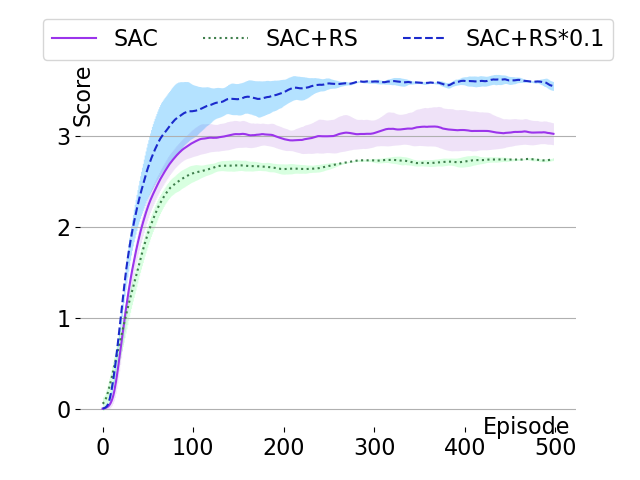}}

    \caption{This figure illustrates the development of the game scores over training episodes where shaded areas correspond to standard deviations. The figure compares the performance of the SAC agents with and without different reward shaping variants: reshaped reward (SAC+RS) and re-scaled reshaped reward combined SAC (SAC+RS*0.1). This figure shows only three examples; more games are presented in additional results \footref{footnote_1} .}
    \label{fig:results_all}
\end{figure*}

Table~\ref{rs-tab:1} shows the comparison of SAC and two variants of reward shaping. The first three games are possible games, and rest of the games are nine difficult games. According to \cite{hausknecht2020interactive}, difficult games tend to have sparser rewards than possible games. Figure \ref{fig:results_all} presents three examples of the game \textit{detective}, \textit{deephome}, and \textit{zork3}.
As Section~\ref{sec:method} outlines, the potential function is based on the soft state value in our experiments. Eight of the twelve games tested scored higher when using a reshaped reward. For the games \textit{deephome} and \textit{zork3} we can see that shaping the original rewards leads to faster convergence than without reward shaping. Thus, we find the reward shaping technique to be particularly advantageous for difficult games as compared to possible games

We compare the re-scaled reshaped reward to the non-scaled reshaped reward. The games \textit{ludicorp} and \textit{balances} have a similar performance for both reward types. However, there is a noticeable gap between the re-scaled reward and the original reshaped reward for games like \textit{jewel}, \textit{zork1}, and \textit{zork3}. To quantify the sparseness of rewards received in each game, we checked the number of steps of each agent before it received the first reward and between two consecutive rewards in the game. The average steps per reward \cite{hausknecht2020interactive} of \textit{zork3} and \textit{karn} are 39 and 17, while for \textit{deephome}, \textit{jewel}, and \textit{zork1} they are 6, 8, and 9. This shows that \textit{zork3} and \textit{karn} have sparser rewards. These results suggest that slightly changed reward values are more beneficial for sparse models.

\section{Limitations and Future Work}
\label{limitations}
As shown in Table \ref{tab:result}, the SAC-based agent improves state of the art on several games, but not all of them. We manually checked the agent-predicted trajectories and the games' walk-throughs. In the following, we discuss examples from the games \textit{balances} and \textit{karn}.

\begin{figure}[t!]
\centering
    \includegraphics[width=\linewidth]{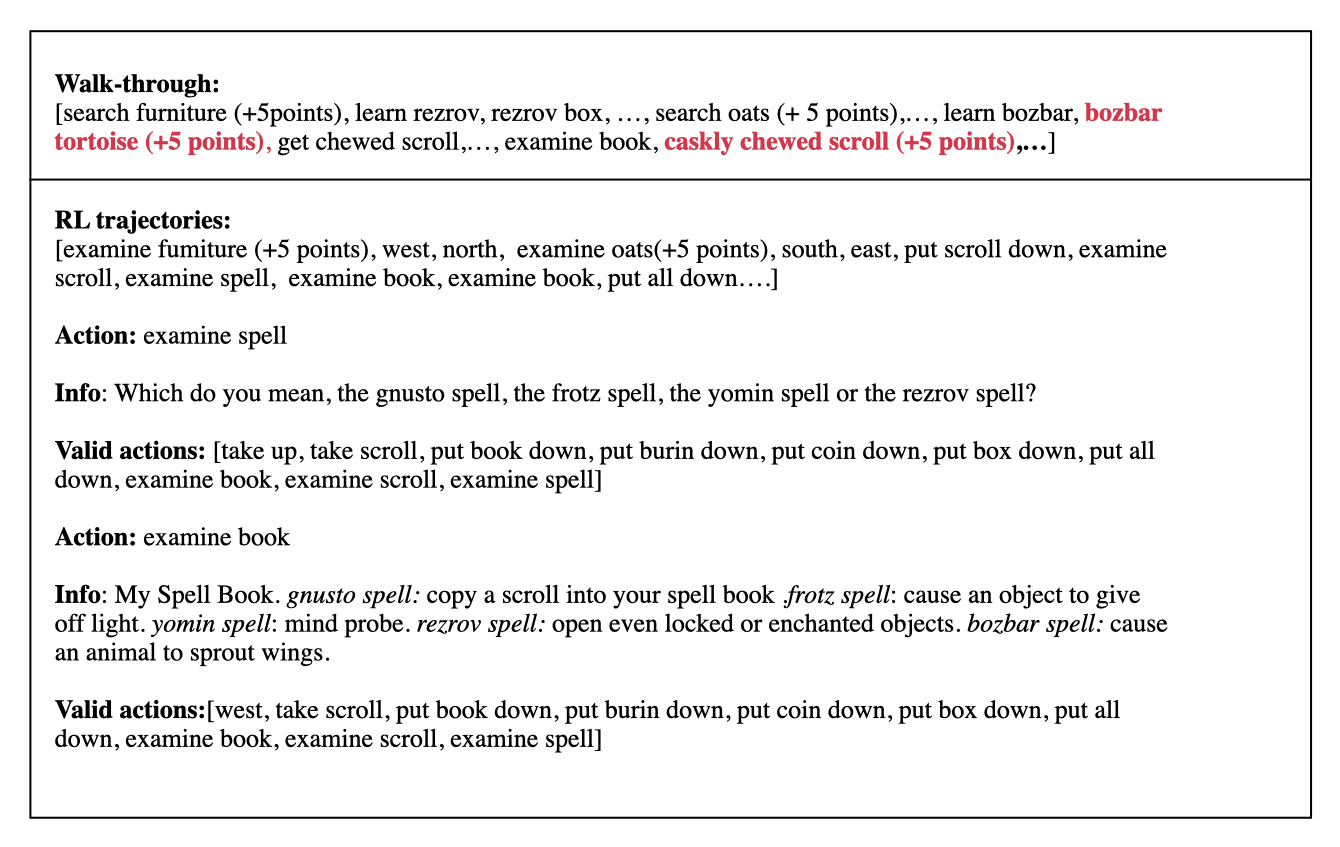}
    \caption{Game balances: The walk-through and RL agent trajectory are shown. The relevant actions, shown in red, are not in the valid action space.}
    \label{fig:limitation}
\end{figure}

The main limitation we found is that the valid action spaces are often incomplete. An essential part of the game \textit{balances} is understanding and using different spells. However, the valid action space does not include those spells, such as \textit{bozbar tortoise} and \textit{caskly chewed scroll}. As shown in Figure~\ref{fig:limitation}, the agent can only repeat meaningless actions and cannot reach higher scores as the required actions, shown in red, are omitted in the valid action space. 
One solution to overcome the imperfection of the valid action space handicap is manually adding some relevant actions from game walk-throughs \cite{tuyls2022multi}. 
Additionally, the agent performs poorly when receiving a large valid action space. For example, compared to \textit{ludicorp} or \textit{jewel}, the game \textit{karn} often receives action spaces with many possible actions at a state.

\begin{figure}[t!]
\centering
    \includegraphics[width=\linewidth]{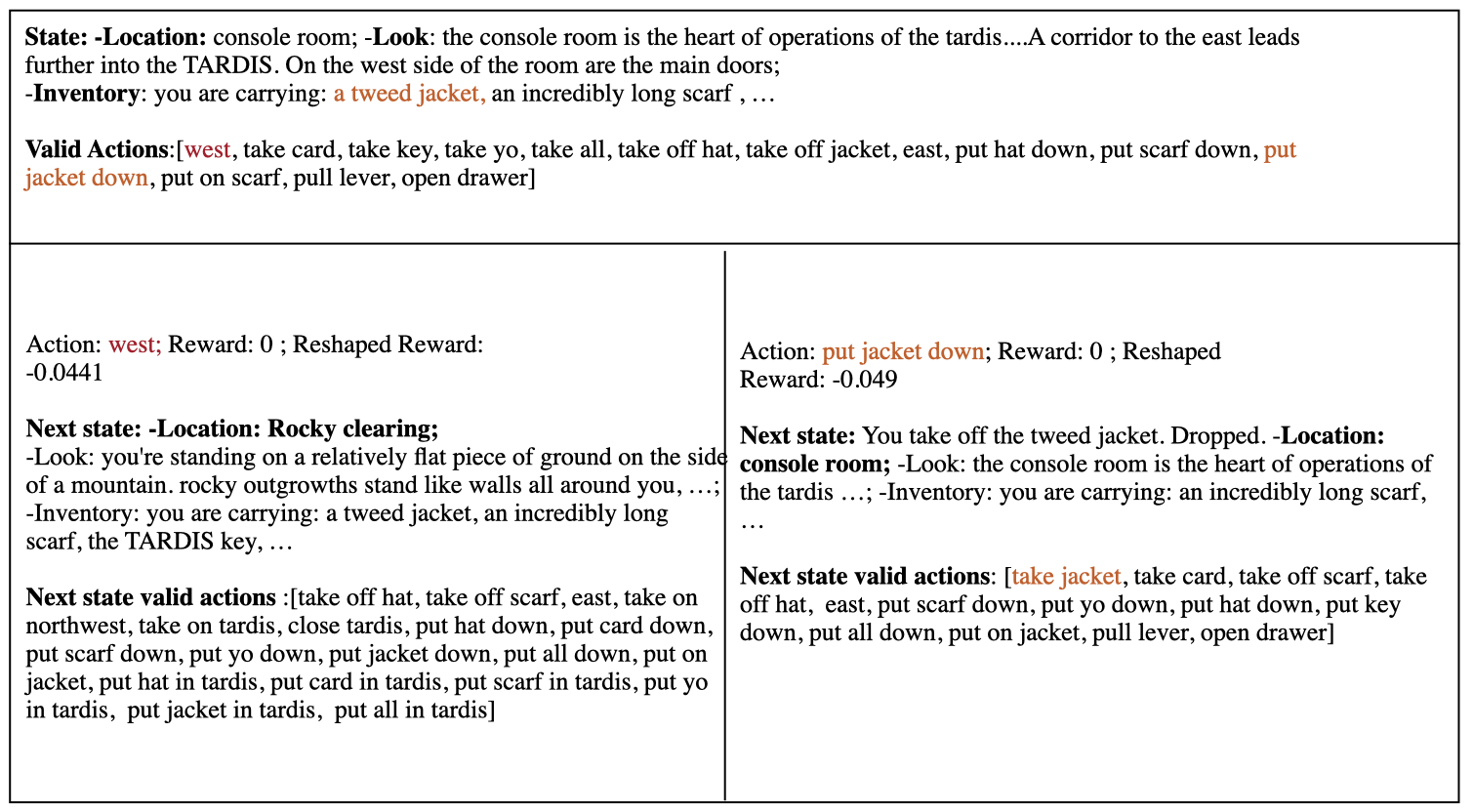}
    \caption{Game \textit{karn}: Most of the actions in the valid action spaces do not lead the agent to a new location and significantly change reward signals. (In the right column, choosing the action \textit{put jacket down}, labeled in yellow, the agent is still in the same location. In the left column, when the agent moves \textit{west}, labeled in red, the agent goes to a new location.) }
    \label{fig:karn_limitation}
\end{figure}

The second limitation is, as Yao \textit{et al.} \cite{yao2021reading} point out, that the current RL agent needs a more robust semantic understanding. Their experiments found that the agent can even achieve higher scores in three games out of twelve using hash-based nonsemantic representations. 
We check the actions predicted by the agent during training as shown in Figure \ref{fig:karn_limitation} to see whether the agent understands the state description. The agent receives the following description:\textit{``Console Room
The console room is the heart of operations of the TARDIS. Dominating the room is a six-side console. Located above the console is the scanner. A corridor to the east leads further into the TARDIS. On the west side of the room are the main doors.''} And the player's inventory includes a jacket, hat, and scarf. The agent gets stuck in the exact location and repeats the same actions: \textit{put jacket down}, \textit{take off jacket}, \textit{take off hat}, and \textit{take card}. The distributions and reshaped rewards change only slightly. The agent tends to require more steps to find practical actions; we assume the agent suffers from making decisions based on semantic understanding. As human players, we can easily decide, like \textit{go east} or \textit{go west}. A further investigation is necessary to ensure the agent learns from language.

In future work, we plan to adapt our method to play with valid actions generated by a large language model (LLM). Action generation is a critical challenge in playing text-based games which requires a high level of language understanding. LLMs are beneficial in generating possible actions \cite{luketina2019survey} since pre-trained LLMs contain rich common sense knowledge. For example, Yao \textit{et al.} \cite{yao-etal-2020-keep} fine-tuned the GPT2 model with game walk-throughs. Nevertheless, it is currently unclear whether an LLM could solve the difficult text-based adventure games we consider in this paper. These games are very difficult to solve, even for intelligent humans, and require a high level of common sense reasoning, creativity, problem-solving capabilities, and planning. Thus, we emphasize that these games pose a challenge to even the best available state-of-the-art language models, and their investigation might help to improve the planning capabilities of future systems of this kind.

In summary, two directions need to be further explored in the future. First, generating accurate action spaces is crucial to improve agent performance. Second, it is necessary to incorporate semantic information in the agent and ensure it is used to predict the next action.

\section{Conclusion}
\label{sec:conclusion}
The primary motivation of this paper is to effectively adapt the maximum entropy RL to the domain of text-based adventure games and speed up the learning process. The results show that the SAC-based agent achieves significantly higher scores than deep Q-learning for some games while using only half the number of training steps. Additionally, we use a reward-shaping technique to deal with sparse rewards. This allows us to learn intermediate rewards, which speeds up learning at the beginning of training for some games and leads to higher scores than without reward shaping for many games. Our analysis reveals two key limitations involving the valid action space that will be addressed in future work.\\

\subsubsection{Acknowledgements} The first author was funded by the German Federal Ministry of Education and Research under grant number 01IS20048. The responsibility for the content of this publication lies with the author. Furthermore, we acknowledge support by the Carl-Zeiss Foundation and the DFG awards BU 4042/2-1 and BU 4042/1-1.

%
%
%
 \bibliographystyle{splncs04}
\bibliography{refs}

\newpage

\section*{Ethical statement}
In this paper, we investigate  RL algorithms in text-based adventure games. We ran our experiments on the Jericho Interactive Fiction game environment, which does not contain any tools for collecting and processing personal data and inferring personal information. Therefore, our experiments do not raise concerns about data privacy. Our RL algorithm could in the future lead to systems that are better at advanced planning and decision making. Such agents would pose many hypothetical dangers. However, we believe that our work does not further any malicious goals or use-cases in itself.

\end{document}